\documentclass[letterpaper, 10 pt, conference]{ieeeconf}
\usepackage{times}
\usepackage{graphicx}
\usepackage{amsmath,amssymb,amsopn,amstext,amsfonts}
\usepackage{cancel}
\usepackage[space]{cite}
\usepackage{pdfsync}
\usepackage{balance}
\usepackage{color}
\usepackage{mathtools}
\usepackage{algpseudocode}
\usepackage{algorithm} 
\usepackage{bm}

\usepackage{diagbox}
\usepackage{float}
\usepackage{epstopdf}
\usepackage{pifont}
\usepackage{fixltx2e}
\usepackage{amsmath}
\usepackage{multirow}
\usepackage{url}
\usepackage{verbatim}
\makeatletter
\let\NAT@parse\undefined
\makeatother
\usepackage[linkcolor=black,citecolor=black,urlcolor=black,colorlinks=TRUE]{hyperref}
\usepackage{booktabs}
\usepackage{graphicx}
\usepackage{subcaption}

\DeclareGraphicsExtensions{.png,.jpg,.eps,.pdf}
\IEEEoverridecommandlockouts
\begin{document}
	\title{Fast-Tracker: A Robust Aerial System for Tracking Agile \\ Target in Cluttered Environments}
	\author{Zhichao Han*, Ruibin Zhang*, Neng Pan*, Chao Xu, and Fei Gao\thanks{All authors are with the State Key Laboratory of Industrial Control Technology, Institute of Cyber-Systems and Control, Zhejiang University, Hangzhou, 310027, China. {\tt\small \{zhichaohan, ruibin\_zhang, panneng\_zju, cxu, and fgaoaa\}@zju.edu.cn} * equal contributors.}}
	
	\maketitle
	\thispagestyle{empty}
	\pagestyle{empty}
	\begin{abstract}
	This paper proposes a systematic solution that uses an unmanned aerial vehicle (UAV) to aggressively and safely track an agile target.
	The solution properly handles the challenging situations where the intent of the target and the dense environments are unknown to the UAV. 
	Our work is divided into two parts: target motion prediction and tracking trajectory planning. 
	The target motion prediction method utilizes target observations to reliably predict the future motion of the target considering dynamic constraints.
	The tracking trajectory planner follows the traditional hierarchical workflow. 
	A target informed kinodynamic searching method is adopted as the front-end, which heuristically searches for a safe tracking trajectory. 
	The back-end optimizer then refines it into a spatial-temporal optimal and collision-free trajectory.
	The proposed solution is integrated into an onboard quadrotor system.
	We fully test the system in challenging real-world tracking missions. Moreover, benchmark comparisons validate that 
	the proposed method surpasses the cutting-edge methods on time efficiency and tracking effectiveness. 
	\end{abstract}
	
	\IEEEpeerreviewmaketitle
	\section{Introduction}
	\label{sec:introduction}
	
	Autonomous aerial tracking is widely applied in aerial photography, surveillance, and security.
	However, it is still challenging to autonomously track a dynamic target with uncertain intent in unknown cluttered environments. 
	For the sake of safety, it is vital for the UAV to recognize both the target and surrounding obstacles with limited sensors and then plan a feasible tracking trajectory. 
	Moreover, in order to deal with unexpected or even fully dynamic situations, high-frequency re-planning is necessary. 
	The above requirements are difficult to meet simultaneously by the limited onboard computing and sensing resources of the system.

	To address the above issues systematically, we propose an aerial tracking framework for applications where the surrounding environments and the intent of the target are unknown to the UAV. 
	Based on the fact that in most tracking scenarios, the target can be approximated as a rigid body, we can safely argue that the target moves with bounded and continuous velocity and acceleration. 
	In the target motion prediction method, we adopt polynomial regression based on the past target observations. 
	Especially, Bernstein basis polynomial is used to enforce dynamical constraints in the regression method.
	The generated trajectory is extrapolated as the prediction of target future motion. 
	On the other hand, due to the occlusion of obstacles, the limited sensing range, and the uncertainty of the target's intent, it is hard for the UAV always to locate the target. 
	We design a strategy so that the UAV can re-locate the target as soon as possible. 
	\begin{figure}[t]
        \centering
		\begin{subfigure}{0.9\linewidth}
			\centering
            \includegraphics[width=1\linewidth]{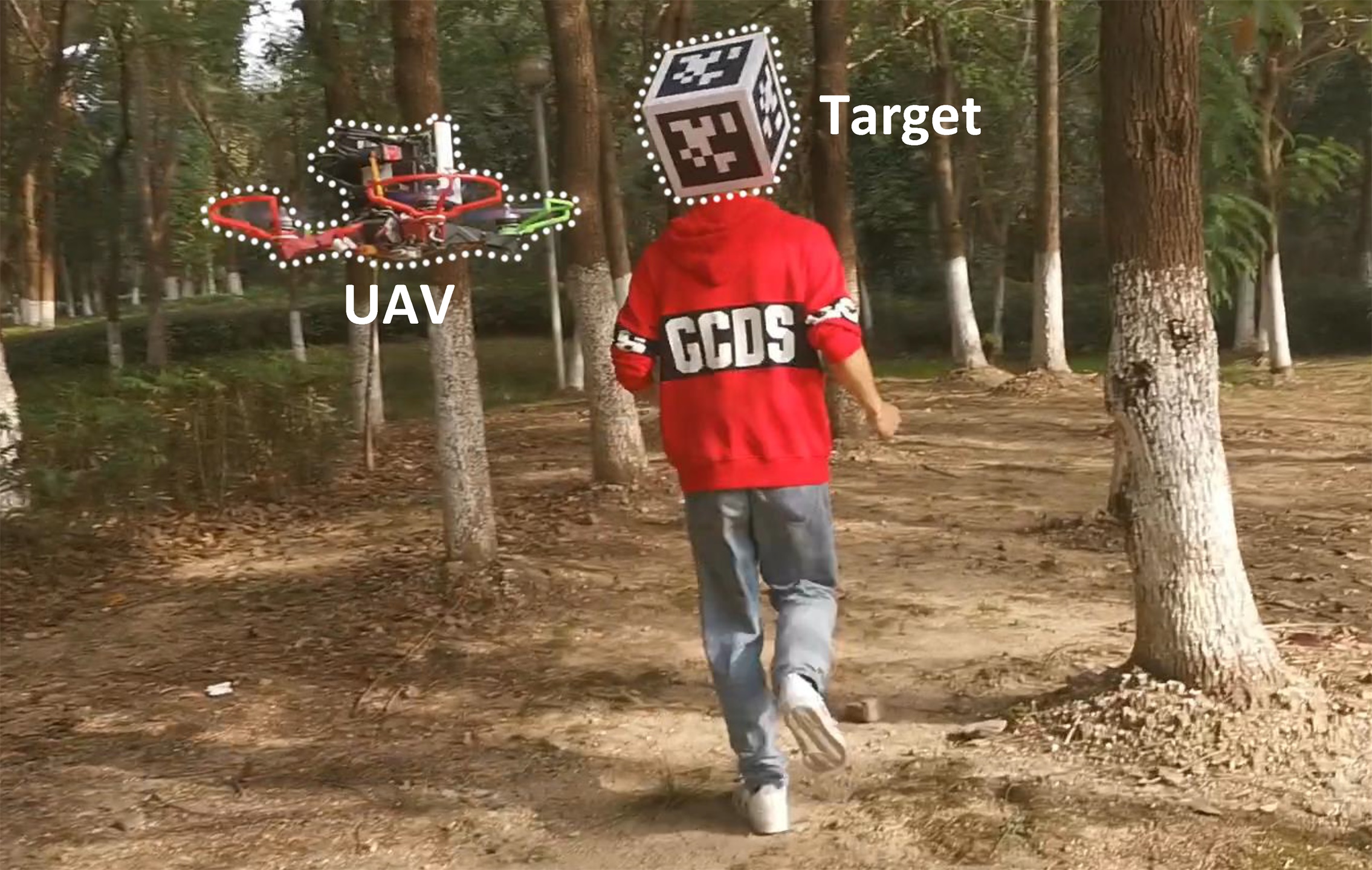}
            \captionsetup{font={small}}
            \caption{The outdoor experiment in a dense forest.}
			\label{pic:top_graph_a}
			\vspace{0.15cm}
        \end{subfigure}
		\begin{subfigure}{0.9\linewidth}
			\centering
            \includegraphics[width=1\linewidth]{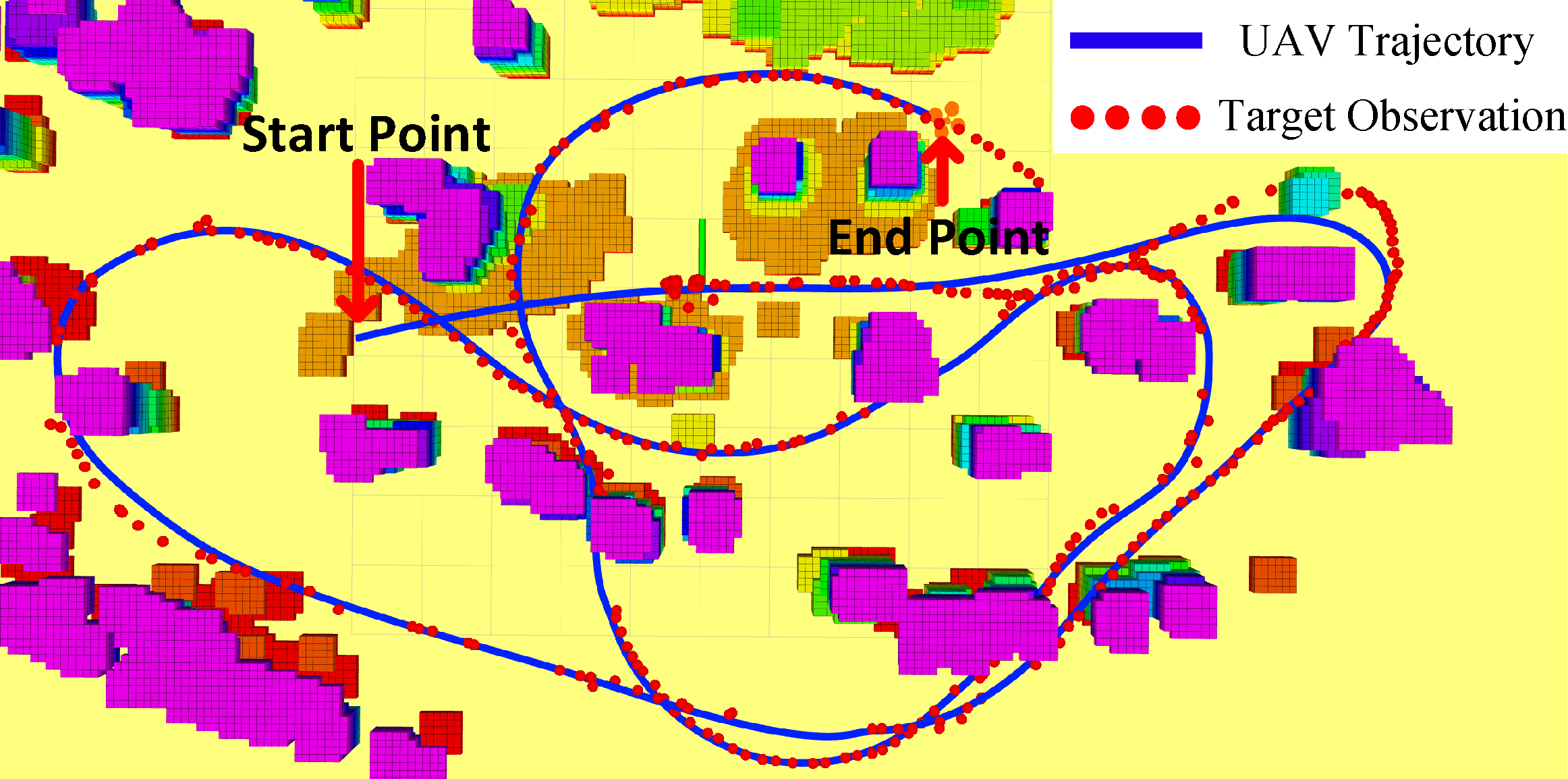}
            \captionsetup{font={small}}
            \caption{Visualization of the outdoor environment, the planned trajectory of the UAV and the target observations.}
            \label{pic:top_graph_b}
		\end{subfigure}
    	\captionsetup{font={small}}
		\caption{
		Aggressive and safe aerial tracking experiment in unknown outdoor cluttered environments.
		}
		\vspace{-1cm}
        \label{pic:top_graph}
	\end{figure}	

\begin{figure}[t]
    	\centering
    	\includegraphics[width=1\linewidth]{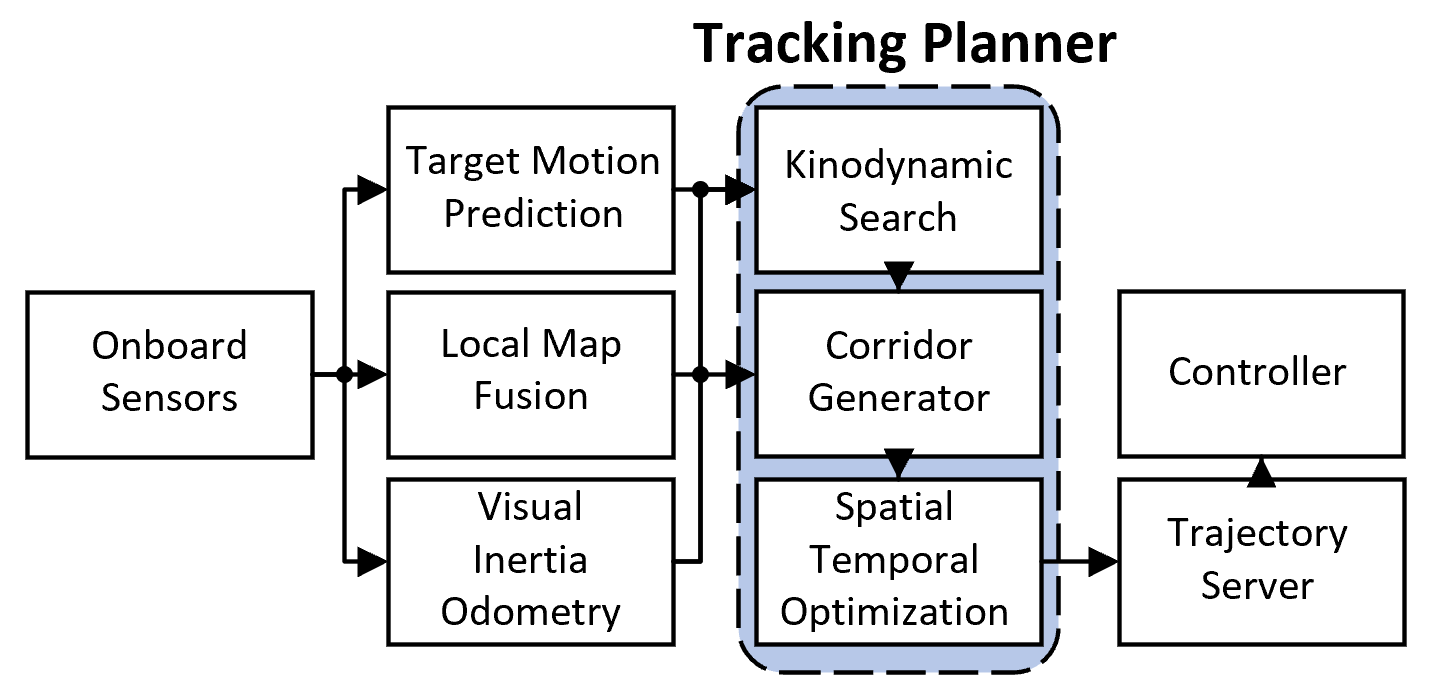}
    	\captionsetup{font={small}}
    	\caption{
			A diagram of the system architecture.
		 }
    	\label{pic: system architecture}
    	\vspace{-0.4cm}
	\end{figure}

	In the tracking trajectory planner, we design a heuristic function for the kinodynamic searching method that considers both the current target observation and motion prediction.
	Afterwards, a flight corridor \cite{JC2016Corridor} that consists of a sequence of connected free-space 3-D grids is formed based on the results of the path searching. 
	The back-end optimizer then generates a spatial-temporal optimal safe trajectory\cite{wang2020DDP} within the flight corridor. 
	Last but not least, we integrate the proposed methods into a customized quadrotor system (as shown in Fig.\ref{pic: system architecture}), with a multi-camera set-up to robustify its sensing capability.
	In this paper, we perform extensive experiments in complex real-world environments to validate that our quadrotor can aggressively and safely track the target.
	We also compare our work with the cutting-edge works in aerial tracking. 
	The results show that our system needs much lower computational overhead and achieves better tracking performance. 
	Contributions of this paper are:
	\begin{itemize}
		\item [1)] 
		A lightweight and intent-free target motion prediction method based on constrained B\' ezier regression.
		\item [2)]
		A safe tracking trajectory planner consisting of a target informed kinodynamic searching front-end and a spatial-temporal optimal trajectory planning back-end. 
		\item [3)]
		The integration of the proposed method with sensing and perception functionality and the presentation of the systematic solution with extensive evaluations.

	\end{itemize}

	\section{Related Work}
	\label{sec:related_works}
	Some previous works \cite{JW2013noc, AG2014noc, HC2017noc} propose vision-based tracking controllers that take the tracking error defined on image space as the feedback. 
	These methods can achieve real-time performance but fail to consider safety constraints, thus limiting the tracking scenarios only in wide-open areas. 
	To incorporate collision avoidance, some methods \cite{TN2017elli, BP2018elli} are proposed. 
	Nägeli et al. \cite{TN2017elli} propose a real-time receding horizon planner that optimizes for visibility to targets and generates collision-free trajectories. 
	Similarly, Penin et al.~\cite{BP2018elli} propose an online replanning strategy based on Model Predictive Control (MPC), which solves a non-linear optimization problem. 
	However, the optimization formulations in \cite{TN2017elli, BP2018elli} are both non-convex, which tends to result in a local minimum.
	They also have a strong assumption on the shape of the obstacle, such as the ellipsoid, which limits the application scenarios to artificial or structural environments.
	As for the target prediction method, similar schemes are proposed in \cite{AG2014noc, HC2017noc, TN2017elli} that estimate the future target motion 
	by Kalman Filter with a coarse motion model which is unreliable.

	Many practical works, such as \cite{Bonatti2018AutonomousDC, BJ2020ICRA, JC2016tracking}, manage to handle tracking scenarios with general environments.
	Bonatti et al. \cite{Bonatti2018AutonomousDC} present a covariant gradient descent based method to find the tracking trajectory by optimizing a set of cost functions considering shot smoothness, obstacle avoidance, and occlusion as well.
	However, the complex cost functions containing non-linear terms are solved depending on the numerical method, which cannot guarantee a satisfactory solution in challenging environments. 
	Jeon et al. \cite{BJ2020ICRA} propose a graph-search based path planner along with a corridor-based tracking trajectory generator.
	This graph-search method builds and traversals a local graph continuously, making it highly time-consuming. 
	Moreover, this method requires a completely known global map of the environment as prior knowledge and assumes the target's intent that it moves through a set of given via-points. 
Therefore, it cannot be used with general unknown environments and targets.
	In \cite{JC2016tracking}, the authors propose an intent-free planner that employs polynomial regression for target motion prediction. 
	A trajectory is then generated by a QP optimizer based on a target prediction obtained by polynomial fitting and exploitation.
	To avoid over-fitting, the polynomial regression method incorporates a regularizer that penalizes the acceleration of the target.  
	Although this prediction method is simple yet effective, it suffers from noisy target observations. 
	Inspired by \cite{JC2016tracking}, we design our prediction module based on polynomial regression, and with several novel extensions to make it generate robust and bounded predictions.

	\section{Intent-free Target Motion Prediction}
	\label{sec:prediction}
	\subsection{Prediction with Constrained B\' ezier Regression}
	\label{sec:bezier_prediction}
	We adopt Bernstein basis polynomial, which is called B\' ezier curve to describe the target predicted trajectory. An n-degree B\' ezier curve is presented as:

	\begin{equation}
		B(t)=\sum_{i=0}^nc_ib_n^i(t),
	\end{equation}
	where each $b_n^i$ is an n-degree Bernstein polynomial basis (described in \cite{gao2018online}), and $[c_0, c_1, ..., c_n]$ is the set of control points of the 
	B\' ezier curve. 
	\par
\begin{figure}[t]
    	\centering
    	\includegraphics[width=1.0\linewidth]{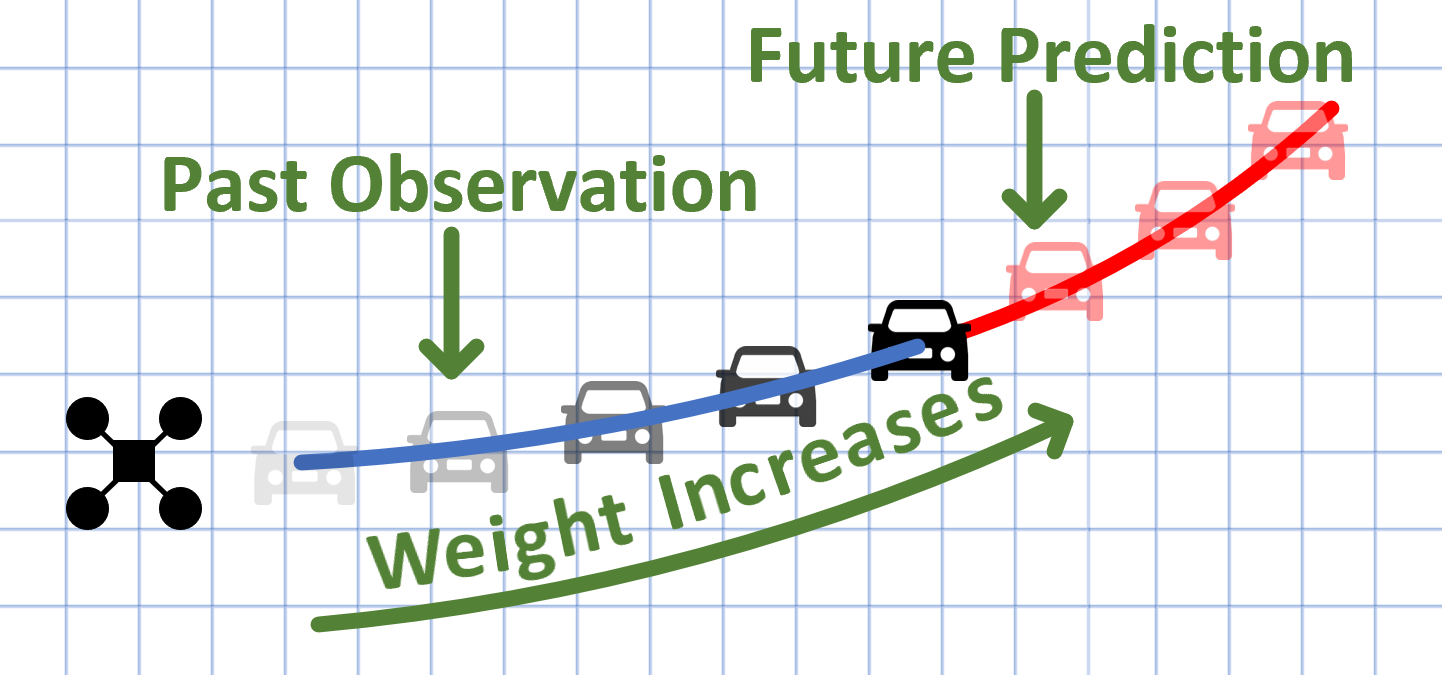}
    	\captionsetup{font={small}}
    	\caption{
			A diagram of the proposed target motion prediction method.
		 }
    	\label{pic: diagram of target prediction}
    	\vspace{-1.1cm}
	\end{figure}

	We denote the observed 3D position of the target in the global frame at time $t \in \mathbb{R}$ as
	$\mathbf{p}_t \in \mathbb{R}^3$. Then a FIFO queue with length $\mathbf{L}$ is maintained, storing the past observations and corresponding timestamp. This queue is denoted
	as $\mathbf{Q}_{target}=[q_1, q_2, ..., q_{L}]$, where $q_i = \{\mathbf{p}_{t_i}, t_i\}$. The time horizon that $\mathbf{Q}_{target}$ includes is $[t_1, t_{L}]$, where $t_{L}$ 
	equals to the current time. When a new target observation is obtained, a new target predicted trajectory $\hat{B}(t)$ is generated as well by fitting the past observations. 
	The trajectory is extrapolated to $(t_{L}, t_{p}]$, during which we predict the target motion. 
	The proposed target motion prediction method is shown in Fig.\ref{pic: diagram of target prediction}.
	\par
	Over time, the confidence of past observations decreases. Therefore, the older an observation is, the smaller the corresponding weight should be in the cost function. 
	We thus add a weight term $w_{t_i}$ to distinguish the confidence of observations with different timestamps. We choose hyperbolic tangent function $tanh(x)$ to compute $w_{t_i}$:
	\begin{equation}
		w_{t_i}=f(t_i)=\left\{
		\begin{aligned}
		& tanh(\frac{k_t}{t_L-t_i}), 					& (i = 1, 2, ..., L-1) \\
		& 1,  					& (i = L)\\
		\end{aligned}
		\right.
		\label{equ:tanh}
	\end{equation}
	whose function value decreases rapidly when the time difference between $t_i$ and the current time $t_L$ grows,
	so that the confidence of different observations can be effectively distinguished.
	\par
	The whole cost function is designed as follows: 
	\begin{equation}
		J_{pre}=\underbrace{\sum_{i=1}^Lw_{t_i}{||\hat{B}(t_i)}-\mathbf{p}_{t_i}||_2^2}_{residual} \  + \  w_{p}\underbrace{L\int_{t_1}^{t_p}{||\hat{B}^{(2)}(t)||_2^2dt}}_{regularizer},
	\end{equation}
	which is to minimize the distance residual between the target trajectory and the observations. A 2-order regularizer is added to avoid over-fitting. $w_p$ represents the weight. 
	\par

	To ensure the dynamical feasibility of the target predicted trajectory, the predicted velocity and 
	acceleration are constrained in $[-v_{mp}, v_{mp}]$ and $[-a_{mp}, a_{mp}]$. Due to the convex hull property and hodograph property of the B\' ezier curve (described in \cite{gao2018online}),
	we just need to enforce the following constraints on each control point $c_i$ in one dimension $\mu$ out of $x,y,z$:
	\begin{align}
		&-v_{mp} \leq {n\cdot(c_\mu^i-c_\mu^{i-1})} \leq v_{mp}  ,\\
		& -a_{mp} \leq {n\cdot(n - 1)\cdot(c_\mu^i-2c_\mu^{i-1}+c_\mu^{i-2})/s_t} \leq a_{mp}, \nonumber
	\end{align}   
	where $n$ is the degree of the B\' ezier curve and $s_t$ is the time scale factor. 
	\par
	The integral $l^2$ norm of the B\' ezier curve has a quadratic form, so the target prediction becomes a constrained QP problem. 
	We adopt Object Oriented software for Quadratic Programming (OOQP)\cite{GM2001OOQP} to build and solve the proposed constrained b\' ezier regression problem. 
	\par

	\begin{figure*}[t]
        \centering
        \begin{subfigure}{0.245\linewidth}
            \includegraphics[width=1\linewidth]{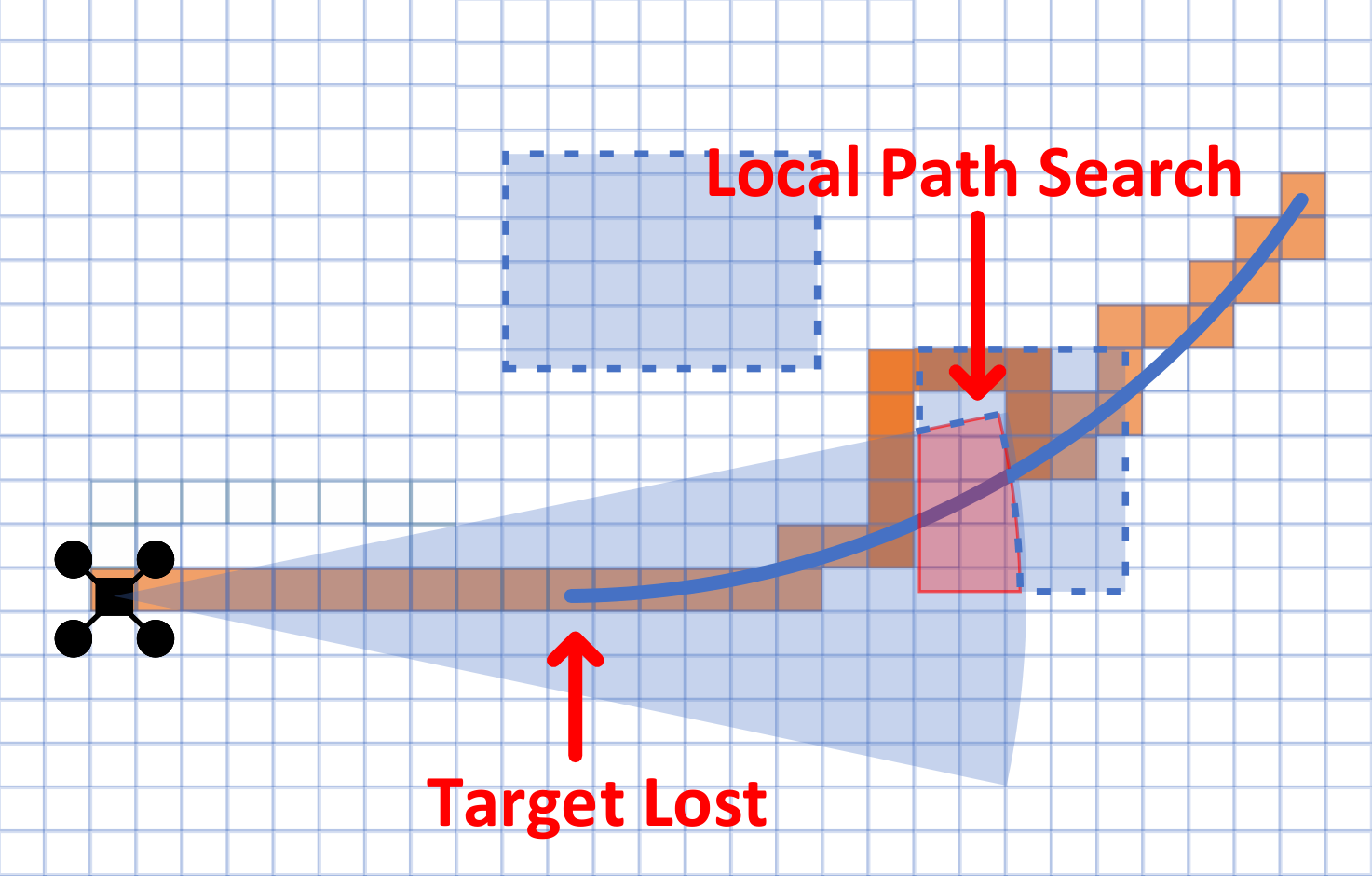}
            \captionsetup{font={small}}
            \caption{Path searching.}
            \label{pic:predict_occ_a}
        \end{subfigure}
        \begin{subfigure}{0.245\linewidth}
            \includegraphics[width=1\linewidth]{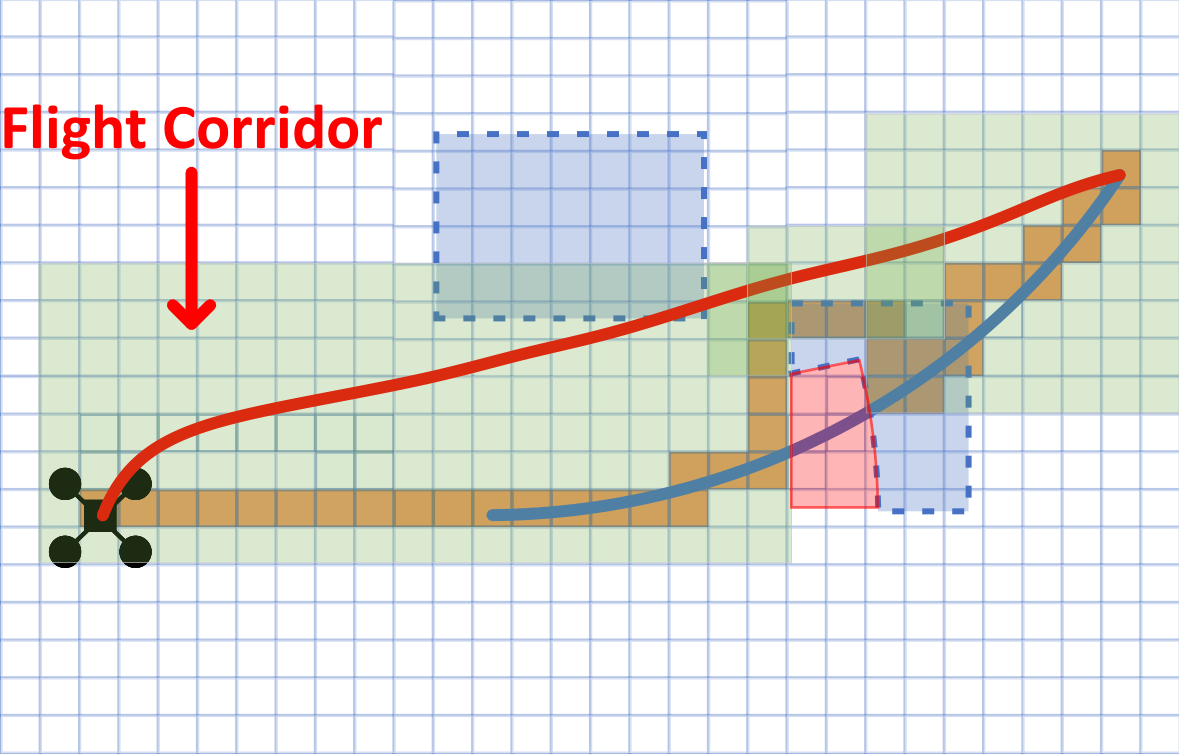}
            \captionsetup{font={small}}
            \caption{Trajectory generation.}
            \label{pic:predict_occ_b}
		\end{subfigure}
        \begin{subfigure}{0.245\linewidth}
            \includegraphics[width=1\linewidth]{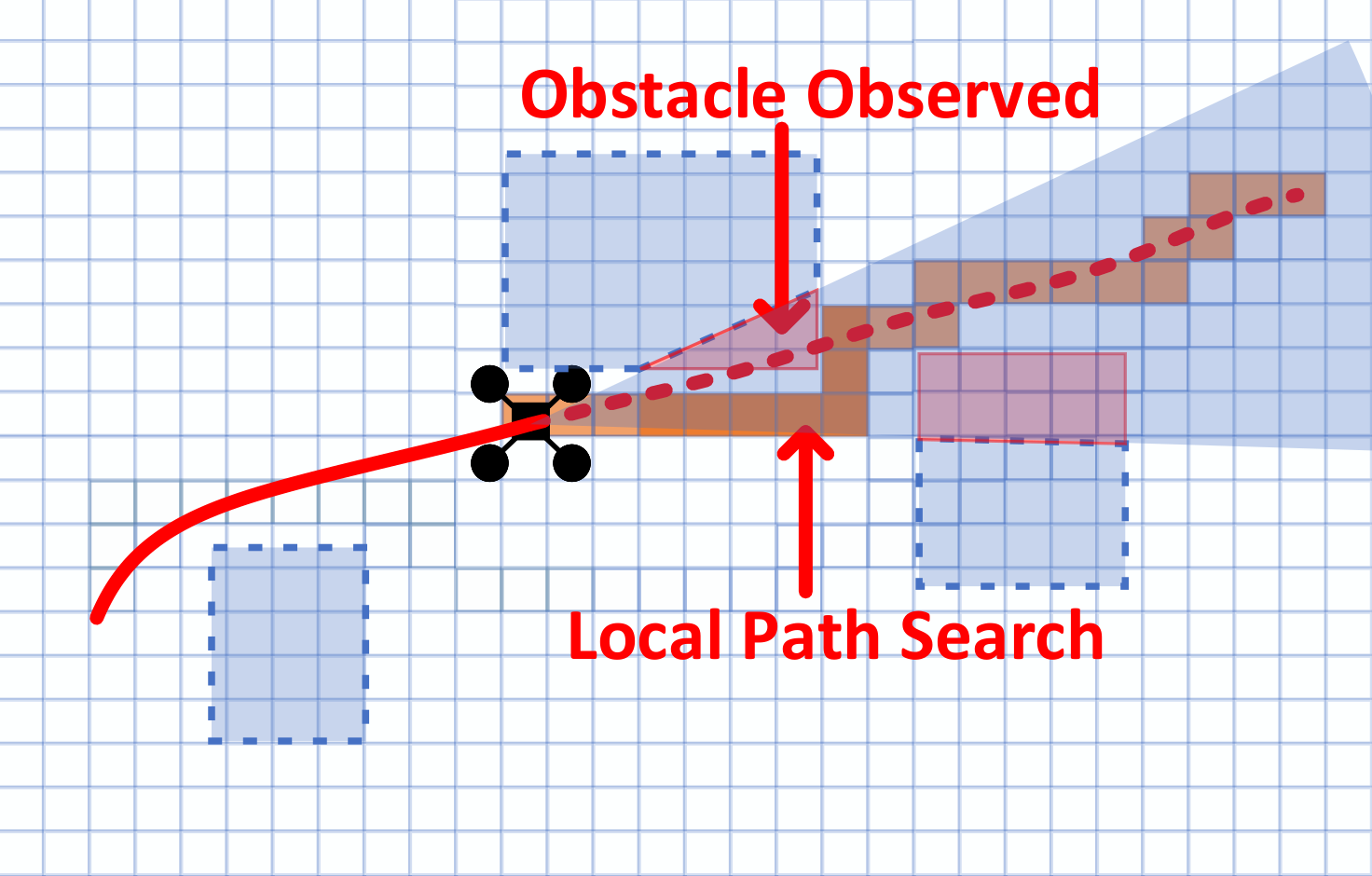}
            \captionsetup{font={small}}
            \caption{Re-planning triggered.}
            \label{pic:predict_occ_c}	
		\end{subfigure}
		\begin{subfigure}{0.245\linewidth}
            \includegraphics[width=1\linewidth]{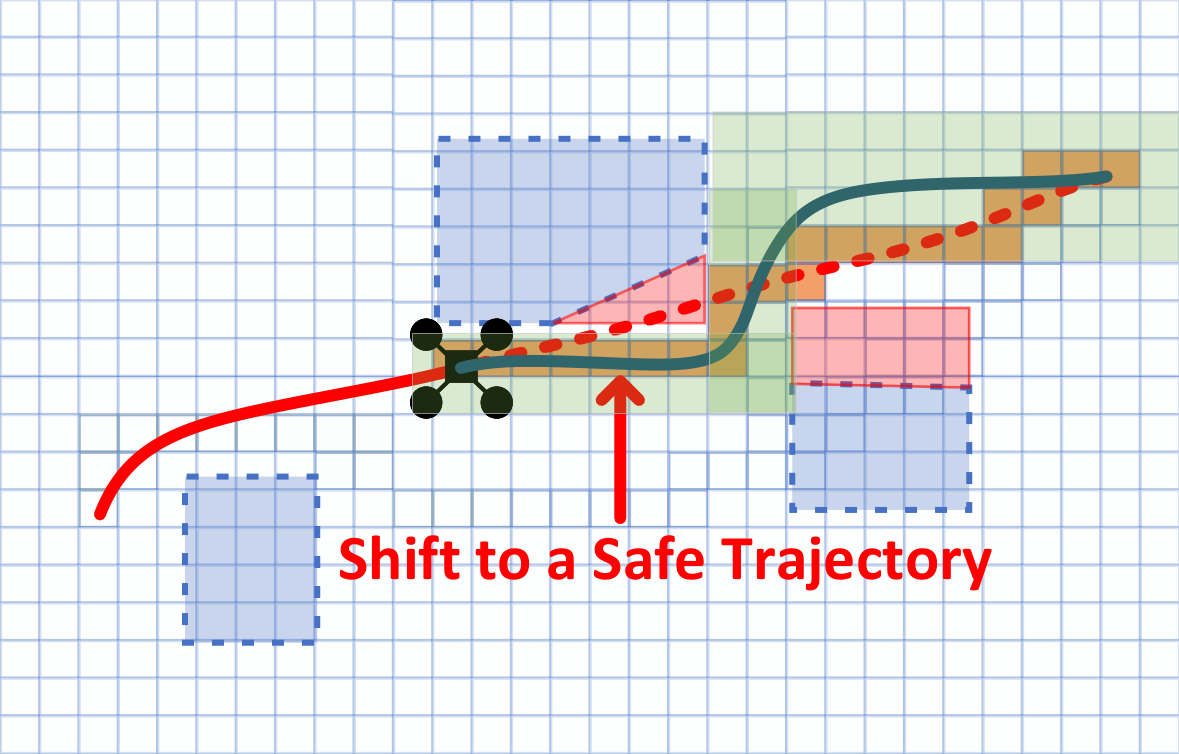}
            \captionsetup{font={small}}
            \caption{Trajectory shifting.}
            \label{pic:predict_occ_d}	
        \end{subfigure}
    	\captionsetup{font={small}}
		\caption{a) $p_{r}$ is generated for the first step. The blue curve is the last target predicted trajectory, and the connected orange blocks make up $p_{r}$.
					When the predicted trajectory collides into an obstacle, the local path search is triggered.
				 b) A flight corridor is generated based on $p_{r}$, then the final re-locating trajectory is generated within the corridor.
				 c) When the UAV is moving along the re-locating trajectory, new obstacles are observed, then the local path search is triggered again.
				 d) A new flight corridor and re-locating trajectory are generated, and the UAV shifts to the new trajectory to ensure safety.
				}
        \label{pic:predict_occ}
	\end{figure*}

	\subsection{Target Re-Locating}
	\label{sec:re-locating}
	As elaborated in Sect.~\ref{sec:introduction}, it is essential to apply an efficient strategy to make the UAV actively explore 
	and re-locate the target if the UAV loses the target observations.
	In most cases, the UAV can re-locate the target by quickly reaching the last target observation and then 
	following the target predicted trajectory. The detailed procedures are as follows:
	\begin{itemize}
		\item [1)] 
		Chiefly, the shortest safe path $p_{obs}$ between the origin (the current position of the UAV) and the 
		destination (the last target observation) is obtained by path searching algorithm. Then, we discretize the last target predicted trajectory into a grid path. 
		If the predicted trajectory collides into obstacles, we use the local path search algorithm to generate a collision-free grid path $p_{pre}$. 
		Afterwards, $p_{obs}$ and $p_{pre}$ is connected as a complete re-locating path $p_{r}$.
		\item [2)]
		A flight corridor is formed based on $p_{r}$. 
		\item [3)]
		A spatial-temporal optimal and collision-free trajectory (described in Sect.~\ref{sec:backend}) is generated within the flight corridor as the final re-locating trajectory.
		\item [4)]
		During the re-locating process, if new obstacles are observed along the re-locating trajectory, a new safe trajectory will be generated.
	\end{itemize}	
	\par
	The target re-locating strategy is illustrated in Fig.\ref{pic:predict_occ}.

	\section{Safe Tracking Trajectory Planning}
	\subsection{Target Informed Kinodynamic Tracking Path Searching} 
	\label{sec:planning}
\begin{figure}[t]
    	\centering
    	\includegraphics[width=1\linewidth]{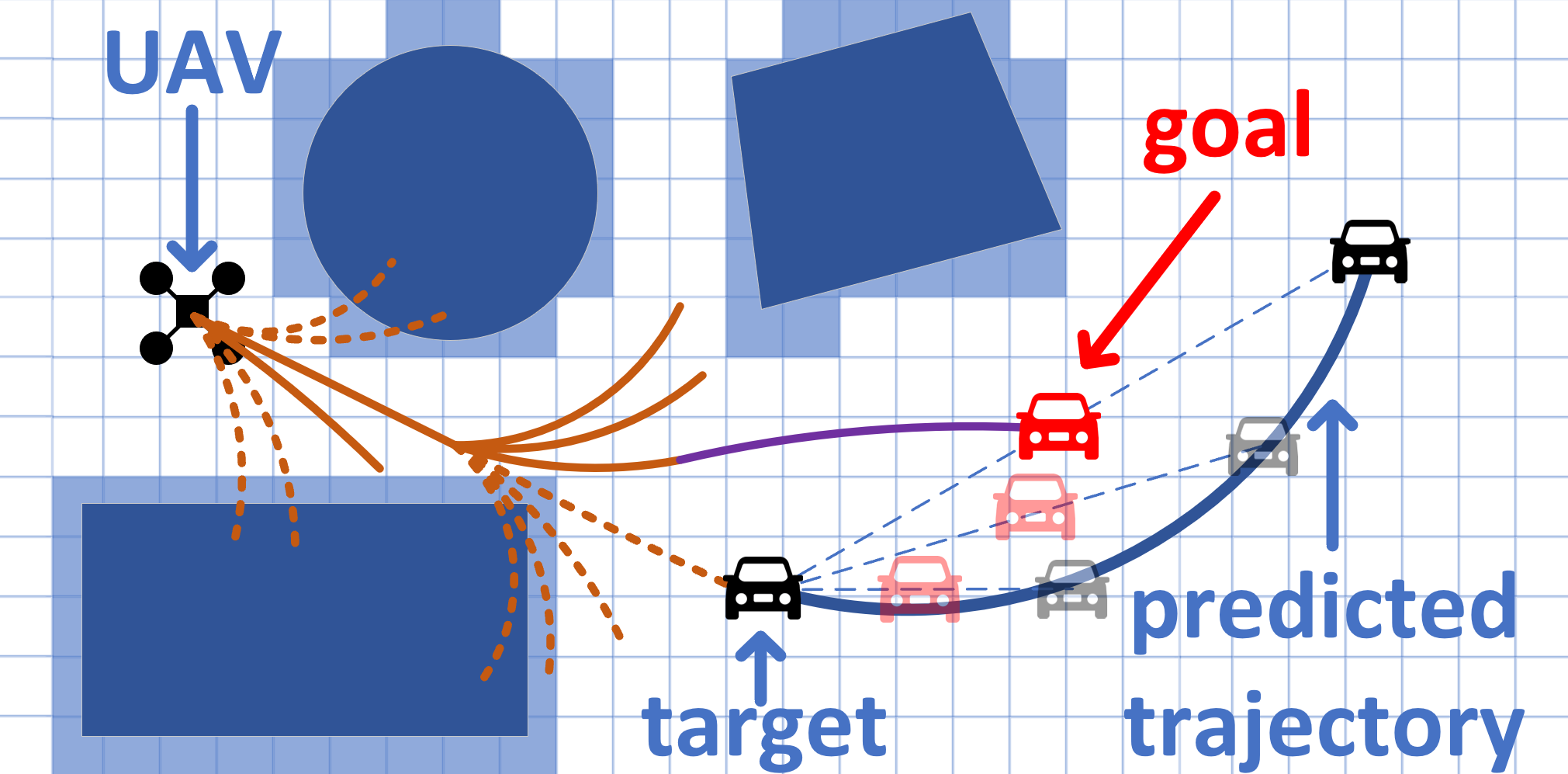}
    	\captionsetup{font={small}}
    	\caption{
			An illustration of the target informed kinodynamic tracking path searcher. 
			The brown curve indicates the expansion process based on Equ.~\ref{equ:Nodes Expansion}, while the purple curve represents the analytic expansion scheme\cite{zhou2019robust} which speeds up searching.
			The navy blue curve stands for the target predicted trajectory. 
			Importantly, the goal state is propagating forwardly during the searching process. This scheme will be explained in Sect.~\ref{sec:Cost Function}.
		 }
    	\label{pic:framework of searcher}
    	\vspace{0cm}
	\end{figure}

	The kinodynamic searching method is based on hybrid-state A$^*$ algorithm\cite{DD2008HybridAstar} that expands nodes (motion primitives) generated by discretizing the control input
	to search for a safe and dynamically feasible trajectory in a voxel grid map.
	We design a heuristic function that fully utilizes the target predicted trajectory generated in Sect.~\ref{sec:bezier_prediction} to quickly search for a target tracking trajectory.
	The whole algorithm framework is shown in Alg.\ref{alg:path finding} and Fig.\ref{pic:framework of searcher}.
	
	\subsubsection{Nodes Expansion}
	\
	\par   
	We use a vector to represent the state of UAV, which is denoted as $\mathbf{x}={(p_{x},p_{y},p_{z},v_{x},v_{y},v_z)}^{T}$.
	The acceleration is taken as the control input $\mathbf{u}$. Let $\mathbf{u} \in \mathcal{U} := [-a_{mt},a_{mt}]^3 \subset \mathbb{R}^3$. 
	We uniformly discretize $\mathbf{u}$ as $\mathbf{u}_D:= \left\{ -a_{mt}, -\frac{n_a-1}{n_a}a_{mt},\cdots,\frac{n_a-1}{n_a}a_{mt}, a_{mt}  \right\}$ in each dimension. 
	Then, the expansion duration is denoted as $\Delta T :=  \left\{\frac{1}{n_t}{\Delta T_m},\frac{2}{n_t}{\Delta T_m}\cdots,\frac{n_t-1}{n_t}{\Delta T_m},{\Delta T_m}  \right\}$. 
	The state transition equation is written as follows:
	\begin{equation}
	\label{equ:Nodes Expansion}
		\mathbf{x_k}=\begin{bmatrix} 1 & \Delta T \\ 0 & 1 \\\end{bmatrix}\mathbf{x_{k-1}} + \frac{1}{2}\begin{bmatrix} \Delta T \\ {2 \Delta T} \end{bmatrix}{\mathbf{u}_D},
	\end{equation}
	where $\mathbf{x_{k-1}}$ is the previous state. Given $\mathbf{x_{k-1}}$, $\mathbf{u}_D$, and $\Delta T$, we can expand the nodes to generate motion primitives.
	Up to $(2n_a+1)^3\cdot n_t$ motion primitives can be generated during one expansion process.
	\par

	\setlength{\textfloatsep}{0pt}
    \begin{algorithm}[t]
        \caption{Kinodynamic Searching For Tracking}
        \label{alg:path finding}
        \begin{algorithmic}[1]
			\State \textbf{Notation}:
			openlist $\Omega_o$, closelist $\Omega_c$, current expansion node $n_c$,
			target predicted trajectory $\hat{B}(t)$, initial state $\mathbf{x_0}$,
		    goal state $\mathbf{x_g}$
			\State Input:  $\hat{B}(t)$, $\mathbf{x_0}$\\
			Initialize()
			\While{$\Omega_o$ is not empty}
			\State $n_c\leftarrow$ FindMinCostNode($\Omega_o$)
			\State $\mathbf{x_g}\leftarrow $  GenerateGoal($n_c$,$\hat{B}(t)$)
			\If{Reach($n_c$,$\mathbf{x_g}$) or AnalyticExpand($n_c$,$\mathbf{x_g}$) }
			\State return OptimalSearchPath()
			\EndIf 
			\State $\Omega_c$.push\_back($n_c$)
			\State $nodes$ $\leftarrow$ Expand($n_c$)
			\For{$n_i$ in $nodes$}
			\State $\mathbf{x_g} \leftarrow $  GenerateGoal($n_i$,$\hat{B}(t)$)
			\If{Nofeasible($n_c$,$n_i$) or $n_i \in \mathbf{\Omega_c}$} 
				\State continue;				
			\EndIf
			\State $g_0\leftarrow n_c.g$+EdgeCost($n_c$,$n_i$)
			\If{$n_i \notin \mathbf{\Omega_o}$} 
			\State $\mathbf{\Omega_o}$.push\_back($n_i$)
			\ElsIf{$g_0>n_i.g$} 
			\State continue
			\EndIf
			\State $n_i.parent\leftarrow n_c$
			\State $n_i.g \leftarrow g_0$
			\State $n_i.f \leftarrow n_i.g+$Heuristic($n_i,\mathbf{x_g}$)
			\EndFor
			\EndWhile
		
        \end{algorithmic}
	\end{algorithm}
	\subsubsection{Cost Function}
	\label{sec:Cost Function}
	\ 
	\par
	The cost function of a node is represented to be $f_c=g_c+h_c$.
	In this formula, $g_c$ represents the actual cost from the initial state $\mathbf{x_0}$ to the current state 
	$\mathbf{x_c}$, while $h_c$ means the heuristic cost to make the search faster.
	We present the actual cost $g_c$ first.
	\par
	To trade off between the control-effort and the time of a trajectory, we minimize its energy-time cost defined as:
	\begin{equation}
		J_{t}(T) =\int_{0}^{T}{||\mathbf{u}(\tau)||}^2d\tau+\mathbf{\rho}T,
	\end{equation}

	Thus, in each expansion process, the cost of a motion primitive generated with the discretized input $\mathbf{u}_D$
	and  $\Delta T$ is  $e_c =(||\mathbf{u}_D||^2 +\mathbf{\rho}){\Delta T} $.
	Based on this definition, if the optimal path $\mathbf{x_0}$ to $\mathbf{x_c}$ consists $m$ motion primitives, $g_c$ can be calculated as: 
	\begin{equation}
		g_c = \sum_{i=1}^{m}e_i
	=\sum_{i=1}^{m}(||\mathbf{u}_{Di}||^2 +\mathbf{\rho}){\Delta T_i}.
	\end{equation}
	\par
	The heuristic function $h_c$ is divided into two parts:
	\begin{equation}
		h_c = D(\mathbf{x_c},\mathbf{x_g})+K_t(S_t-t_c).
	\end{equation}
	\par
	The first term $D(\mathbf{x_c},\mathbf{x_g})$ is the distance between the current state of the UAV
	$\mathbf{x_c}$ and the goal state $\mathbf{x_g}$. 
	To make the searching result prospective, we do not directly use the current state of the UAV as the goal state $\mathbf{x_g}$,
	but use the weighted value of $\mathbf{x_{tc}}$ and $\mathbf{x_{tp}}$ instead. 
 	$\mathbf{x_g}$ is computed as follows:
	\begin{equation}
		\mathbf{x_g} = (1-\mathbf{\phi})\mathbf{x_{tc}}+\mathbf{\phi}\mathbf{x_{tp}},
	\end{equation}
	where $\mathbf{\phi}$ is the weight. As the path expansion time $\mathbf{\tau}$ increases, $\mathbf{x_{tp}}$
	propagates along $\hat{B}(t)$. We synchronize the time axis of the path expansion process and the 
	target predicted trajectory. Specifically, at time $\tau$ of path expansion, $\mathbf{x_{tp}(\tau)} = \{\mathbf{p}_{tp}, \mathbf{v}_{tp}\} =\{\hat{B}(\tau), \hat{B}^{(1)}(\tau)\}$.
	\par
	
\begin{figure}[t]
    	\centering
    	\includegraphics[width=0.9\linewidth]{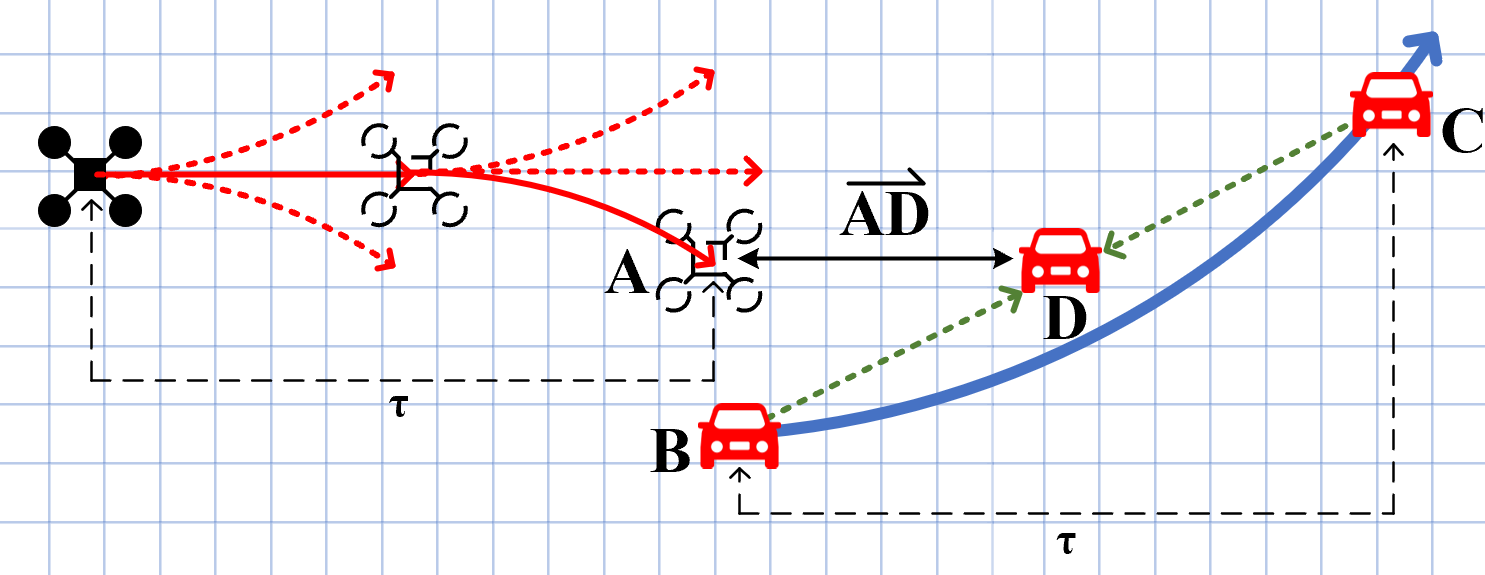}
    	\captionsetup{font={small}}
    	\caption{
			An illustration of $D(\mathbf{x_c},\mathbf{x_g})$. $\mathbf{A}$ represents $\mathbf{x_c}$ when the expand time is $\tau$.
			$\mathbf{B}$ represents $\mathbf{x_{tc}}$, the current state of the target. $\mathbf{C}$ stands for $\mathbf{x_{tp}}$, the predicted state at time $\tau$. $\mathbf{D}$ is $\mathbf{x_g}$, the weighted state of $\mathbf{x_{tc}}$ and $\mathbf{x_{tp}}$. 
			$D(\mathbf{x_c},\mathbf{x_g})$ is computed as the OBVP distance between $\mathbf{A}$ and $\mathbf{D}$.
		 }
    	\label{pic:Astar_heuristic}
    	\vspace{-0.3cm}
    \end{figure}

	An optimal boundary value problem (OBVP) proposed in \cite{zhou2019robust} solves the minimum dynamic cost of an optimal path. 
	We define the minimum cost solved by this problem between $\mathbf{x_c}$ and $\mathbf{x_g}$ as the OBVP distance, and take it as $D(\mathbf{x_c},\mathbf{x_g})$. The computation of $D(\mathbf{x_c},\mathbf{x_g})$ is illustrated in Fig.\ref{pic:Astar_heuristic}.
	\par

	On the other hand, $K_t(S_t-\tau)$ is a time penalty term, where $S_t$ is the sum of expected expansion time, and $K_t$ is the weight. 
	This term greatly speeds up searching since it tends to choose the current node’s neighboring region instead of the whole 
	state space. We trade off optimality against computing efficiency by adding this term. However, the results (Sect.~\ref{sec:results}) validate that the 
	planner can still find a feasible and satisfactory solution in most situations.

	\subsection{Spatial-Temporal Optimal Trajectory Generation}	
	\label{sec:backend}
	In the back-end, we use the efficient optimization method proposed in \cite{wang2020DDP} to generate a piecewise-polynomial trajectory $p(t)$ which is spatial-temporal optimal
	within a given safe flight corridor.
	The method reduces the decision variables of the involved optimization into intermediate waypoints $q_w$ and piece times $T$ of the piecewise-polynomial trajectory.
	\par
	Chiefly, a flight corridor $\mathcal{F}$ based on the trajectory obtained from the front-end is generated, defined as:
	\begin{equation}
		\mathcal{F} = \bigcup\limits_{i=1}^M \mathcal{C}_i,
	\end{equation} 
	where each $\mathcal{C}_i$ is a finite cube:
	\begin{equation}
		\mathcal{C}_i = \{ x \in \mathbb{R}^3  | \mathbf{A}_{c_{i}} x \leq b_{c_{i}} \}.
	\end{equation}
	
	This method takes $\mathcal{F}$ as input and minimizes the cost function as follows:
	\begin{equation}
	J_{\sum}(q_w,T)={J_S}(q_w,T)+J_F(q_w)+J_D(q_w,T).
	\end{equation}
	The smoothness cost ${J_S}(q_w,T)$ in one dimension is described as: 
	\begin{equation}
	J_{S_{\mu}} = \sum_{i=1}^{M} \int_{0}^{T_i}{\left\|{p_{i_{\mu}}^{(3)}(t)}\right\| }^2 dt.
	\end{equation}
	$J_F(q_w)$ is a logarithmic barrier term to ensure that each $q_{w_i}$ is constrained in $\mathcal{C}_{i} \cap \mathcal{C}_{i+1}$,
	defined as:
	\begin{equation}
	J_F(q_w) = -\kappa \sum_{i=1}^{M-1}\sum_{j=i}^{i+1}\mathbf{1}^T \ln \left[ b_{c_{j}} -\mathbf{A}_{c_{j}} q_{w_i}  \right].
	\end{equation}
	where $\kappa$ is a constant coefficient, $\mathbf{1}$ an all-ones vector and $\ln \left[ \cdot \right]$ the entry-wise natural logarithm.
	The last cost term $J_D(q_w,T)$ is a penalty to adjust the aggressiveness of the whole trajectory, and is defined as:
	\begin{align}
		&J_D(q_w,T)=\rho_t\sum_{i=1}^{M}T_i+           \\
		&\rho_v\sum_{i=1}^{M-1}  l( {\left\| \frac{q_{w_{i+1}}-q_{w_{i-1}}}{T_{i+1}+T_{i}} \right\|}^2 - {v_m}^2)+ \nonumber \\
		&\rho_a\sum_{i=1}^{M-1}l({\left\|  \frac{(q_{w_{i+1}}-q_{w_i})/T_{i+1}-(q_{w_i}-q_{w_{i-1}})/T_i}{(T_{i+1}+T_i)/2}   \right\|}^2 -a_m^2) \nonumber
	\end{align}
	where $l(x) = max(x,0)^3$, $v_m$ is the maximum velocity and $a_m$ is the maximum acceleration.
	Morever, a novel splitting procedure is utilized in this method to suppress any possible constraint violation. 

	\begin{figure}[t]
    	\centering
    	\includegraphics[width=0.9\linewidth]{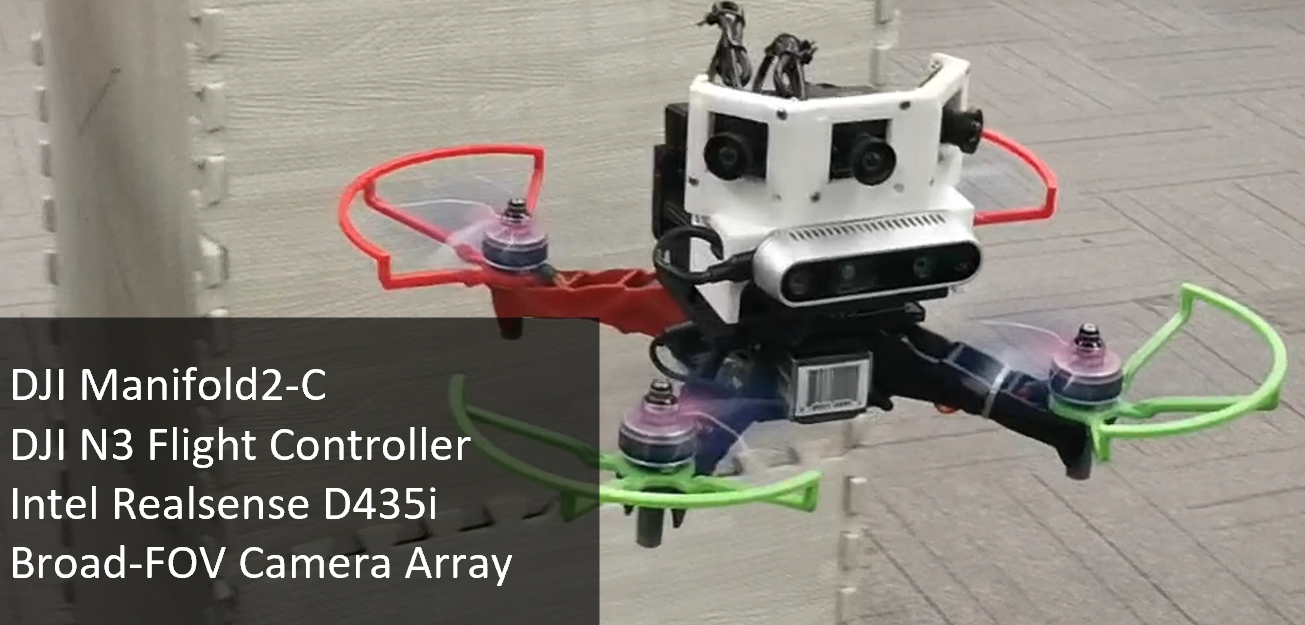}
    	\captionsetup{font={small}}
    	\caption{
			Overview of our quadrotor system.
		 }
    	\label{pic:quadrotor}
    	\vspace{-0.2cm}
	\end{figure}
	
	\section{Results}
	\label{sec:results}
	\subsection{Implementation details}
	\label{sec:implementation details}
	Real-world experiments are presented on the same quadrotor platform of \cite{gao2019teach}. 
	The target is set as a tag carried by a freely moving human. We use AprilTag \cite{EO2011Apriltag}, a visual fiducial system, to detect the tag.
	Furthermore, a broad-FOV (field-of-view) camera array consisting of 3 cameras is implemented to detect the target especially. Credit to the broad FOV for target detection, the quadrotor does not have to face the target. 
	We correspondingly plan the quadrotor to face the tangential direction of the tracking trajectory, ensuring that the surrounding obstacles can be fully observed. 
	The proposed system runs with approximately $20ms$ in total.
	We set the re-planning frequency of the whole system as 15Hz, sufficient for most tracking missions. The quadrotor system of the proposed methods is shown in Fig.\ref{pic:quadrotor}.
	The parameter setting of the proposed methods is given in the attached video. 

	\begin{figure}[t]
        \centering
		\begin{subfigure}{0.9\linewidth}
			\centering
            \includegraphics[width=1\linewidth]{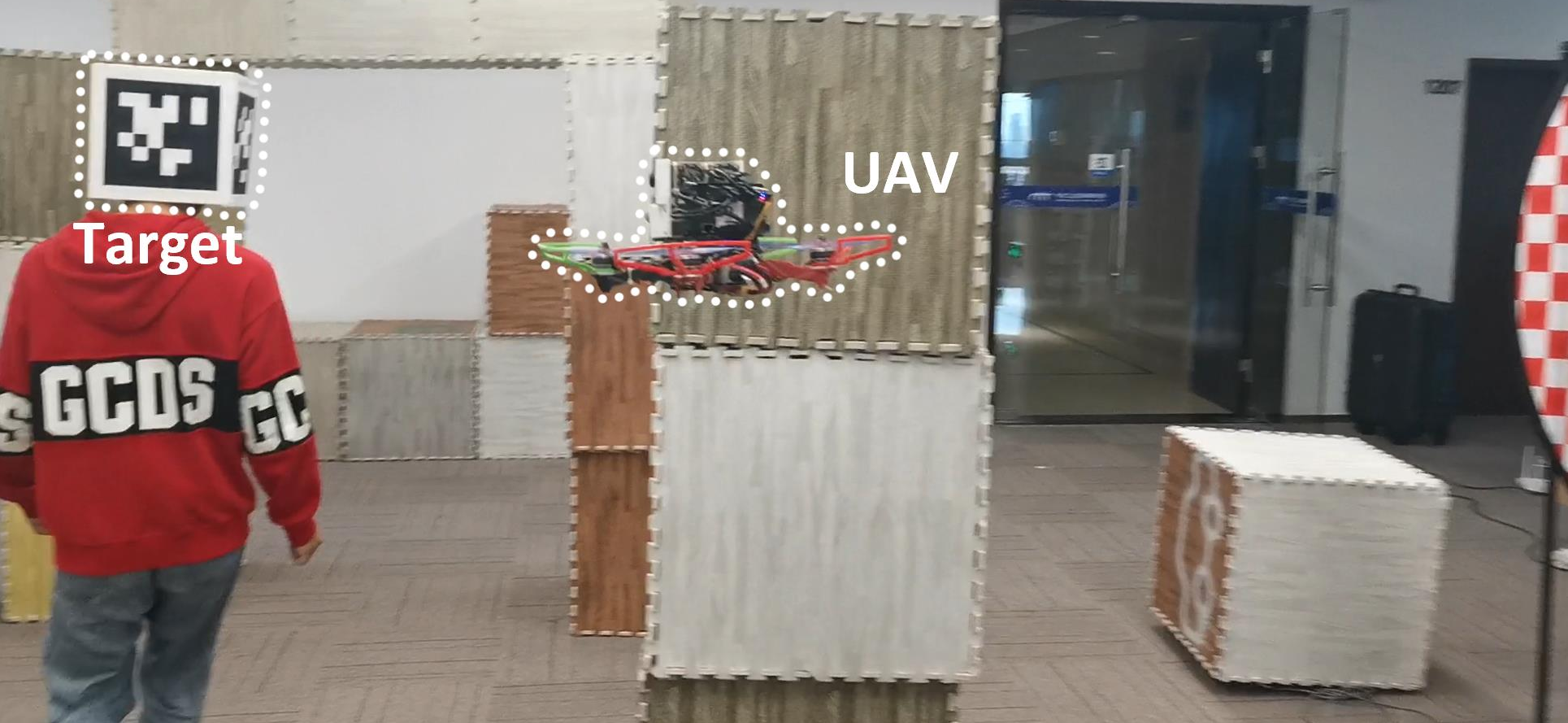}
            \captionsetup{font={small}}
            \caption{Demonstration of the indoor experiment.}
			\label{pic:indoor_env}
			\vspace{0.15cm}
        \end{subfigure}
		\begin{subfigure}{0.9\linewidth}
			\centering
            \includegraphics[width=1\linewidth]{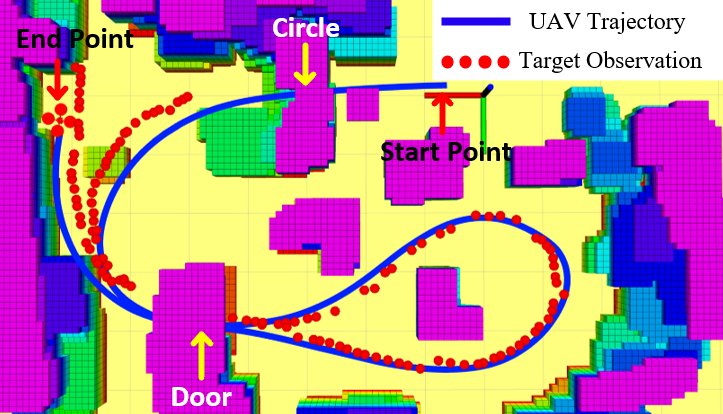}
            \captionsetup{font={small}}
            \caption{Visualization of the indoor environment, the planned trajectory of the UAV and the target observations.}
			\label{pic:indoor_vis}
			\vspace{0cm}
		\end{subfigure}
    	\captionsetup{font={small}}
		\caption{
		Safe aerial tracking experiment in unknown indoor environments with massive obstacles.
		}
		\vspace{-0.2cm}
        \label{pic:indoor}
	\end{figure}

	\subsection{Real-World Experiments}	
	\label{sec:experiment} 
	We present several experiments in unknown in(out)-door cluttered environments. The indoor environment is carried out in a room with many obstacles as shown in Fig.\ref{pic:indoor_env} and Fig.\ref{pic:indoor_vis}.   
	In this experiment, the target moves at an average speed of $1.3m/s$. Fig.\ref{pic:indoor_vis} illustrates that the quadrotor closely follows the target without colliding with obstacles in such a challenging environment.
	Moreover, the tracking trajectory keeps smooth and dynamically feasible during the whole experiment. 
	\par
	In the outdoor environment, the target runs fast in a dense forest, as demonstrated in Fig.\ref{pic:top_graph_a}. This environment further challenges the proposed system. 
	As is shown in Fig.\ref{pic:top_graph_b}, It turns out that the quadrotor still closely follows the fast-moving target, and the tracking trajectory is satisfactory as well. 
	During the tracking mission, the quadrotor reaches $3m/s$ in such a complex environment.
	\par
	We also conduct experiments to especially illustrate the target re-locating strategy and test the proposed system's robustness. More details are available in the attached video.

	\subsection{Benchmark Comparisons}
	\label{sec:benchmark}	
	In this section, we both compare the proposed target motion prediction method and tracking trajectory planner 
	with cutting-edge methods.
	A simulated $20\times20\times3m$ environment generated with 140 randomly deployed obstacles is adopted for benchmark
	comparisons. 
	\par
	Chiefly, we compare our work with \cite{JC2016tracking}. Only the target motion prediction method 
	is compared because the workflow of the planner proposed in \cite{JC2016tracking} varies widely from ours.
	In the comparisons, the ground truth of target position corrupted by Gaussian noise with zero mean is offered as the target observation. The common parameters are selected
	 as the same ($n=5$, $L=30$, $t_p - t_L = 2.5s$, $w_p = 15$). Different standard deviation
	of $0.05m$, $0.3m$, $0.6m$ of the Gaussian noise separately represents low-noise, middle-noise and high-noise environments.
	We compare the average distance error between the target's future position $\mathbf{p}_f(t)$ and the predicted trajectory $\hat{P}(t)$. 
	The average distance error is computed as:
	\begin{equation}
		\sum_{i=1}^{L_f}\frac{{||\hat{P}(t_0 + i \cdot t_s)}-\mathbf{p}_f(t_0 + i \cdot t_s)||_2}{L_f},
	\end{equation}
	where $t_0$ is the current time, $t_s=0.05s$ is the sampling interval, and $L_f=50$ is the sampling size. 
	Over 15000 target motion predictions are done in each environment. 

	\par
	As is shown in Tab.\ref{tab:predict_cmp}, the proposed method is superior in terms of distance error in each
	environment. The results state that the proposed target motion prediction method is more accurate and reliable.
	The computation time of the proposed method is about $0.5ms$, while the method\cite{JC2016tracking} 
	takes up about $0.3ms$. The increase in time consumption is almost negligible compared to the improvement
	in prediction accuracy, validating the practicality of the proposed method.
	\begin{table}[t]
		\centering
		\caption{Target Motion Prediction Method Comparison.}
		\renewcommand\arraystretch{1.1}  
		\begin{tabular}{|c|ccc|}
		\hline
		Method & $\text{error}_\text{low}$(m) & $\text{error}_\text{median}$(m) & $\text{error}_\text{high}$(m)   \\
		\hline
		Proposed & \bf 1.82  & \bf 2.54 & \bf3.45   \\
		\hline
		Method\cite{JC2016tracking} & 3.70 & 4.66 & 5.60   \\
		\hline
		\end{tabular}
		\label{tab:predict_cmp}
		\vspace{0cm}
	\end{table}

	\begin{table}[t]
		\centering
		\caption{Tracking Trajectory Planner Comparison in \\ simple, median and hard scenarios.}
		\setlength{\tabcolsep}{1.4mm}
		\renewcommand\arraystretch{1.2}
		{
		\begin{tabular}{|c|c|c|c|c|c|}
		\hline
		Scenario             & Method       & $t_\text{front}$(ms) & $t_\text{back}$(ms) & $t_\text{total}$(ms) & $r_t$(\%) \\  \hline
		$\bar v=1.2m/s,$ & Proposed      & \bf 6.3        & \bf1.0        & \bf7.3        & \bf98.2             \\ \cline{2-6}
		$v_m=2.3m/s$	& Method\cite{BJ2020ICRA} & 222.7       & 11.3      & 234.0       & 91.4             \\ \hline
		$\bar v=1.5m/s,$ & Proposed      & \bf12.1        & \bf1.7        & \bf13.8        & \bf91.4             \\ \cline{2-6}
		$v_m=2.9m/s$	& Method\cite{BJ2020ICRA} & 240.5       & 11.4       & 251.9       & 85.1             \\ \hline
		$\bar v=2.1m/s,$  & Proposed      & \bf15.8         & \bf2.7          & \bf18.5         & \bf85.6             \\ \cline{2-6}
		$v_m=3.9m/s$	& Method\cite{BJ2020ICRA} & 363.8       & 10.9       & 374.7         & 69.3            \\
		\hline
		\end{tabular}}
		\label{tab:planner_cmp}
		\vspace{-0.9cm}
	\end{table}

	\begin{figure}[t]
    	\centering
    	\includegraphics[width=0.9\linewidth]{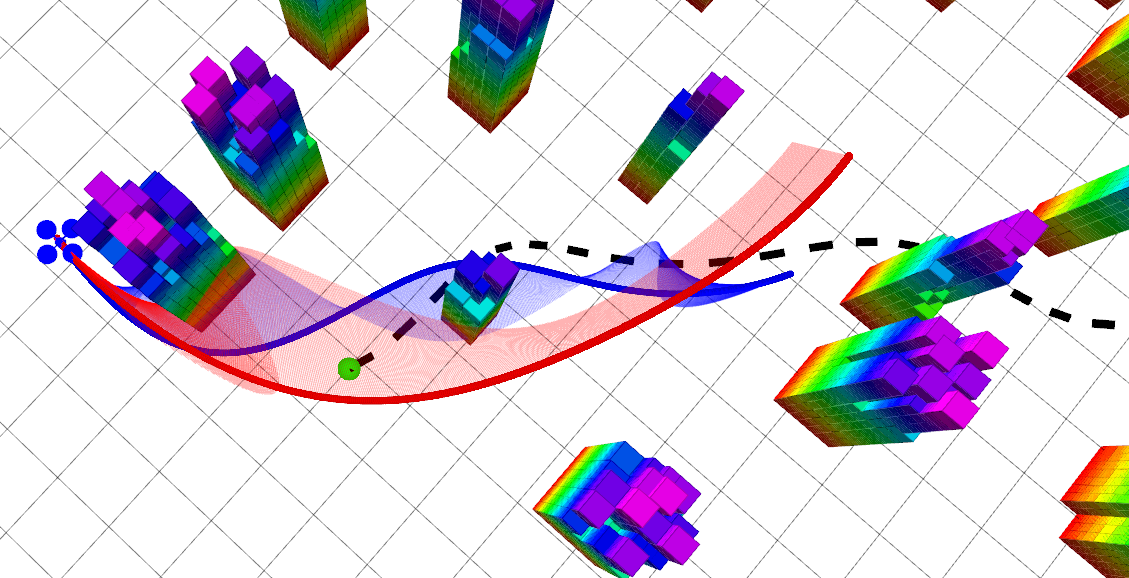}
    	\captionsetup{font={small}}
    	\caption{
			Comparison between the tracking trajectories. The black dash line represents the target's future motion.
			The blue curve and cloud is the tracking trajectory and its acceleration generated by the proposed planner. The red ones are generated
			by planner\cite{BJ2020ICRA}.
		 }
    	\label{pic:benchmark}
    	\vspace{0.3cm}
	\end{figure}

	\par
	Afterwards, we compare our work with \cite{BJ2020ICRA}. Because the target motion prediction method 
	in \cite{BJ2020ICRA} cannot work without a set of given via-points for the target, this method differs greatly from the proposed one.
	Therefore, only the tracking trajectory planner is compared. 
	In each simulated tracking mission, the target randomly moves across the environment. The ground truth of the target's future 
	trajectory is directly provided with the two planners as input. 
	In the tracking mission, we define effective tracking time as the time period when
	the distance between the quadrotor and the target in the $x-y$ plane is smaller than $3m$.
	The proportion of the effective tracking time to the total tracking time is defined as tracking rate $r_t$.
	We further set up different tracking scenarios in which the target has different average speed $\bar v$ and maximum speed $v_{m}$.
	The computation time and $r_t$ are compared in 200 tracking missions for each scenario.
	\par
	From Tab.\ref{tab:planner_cmp}, we first conclude that the proposed planner needs a much lower computation budget. 
	Also, the proposed planner achieves better tracking effectiveness. This is because the proposed planner 
	reacts better to unpredictable cases such as the target's sudden acceleration credit to the high-frequency re-planning.
	Moreover, Fig.\ref{pic:benchmark} shows that the tracking trajectory formed by the proposed planner has lower acceleration.
	Thus, it is more energy-saving for the UAV.
	\vspace{-0.14cm}
	\section{Conclusion}
	\label{sec:conclusion}
	\vspace{-0.14cm}
	In this paper, we propose a systematic solution for aerial tracking. A target motion prediction method for forecasting the future target motion utilizing
	past target observations is proposed as the first phase. We then propose a heuristic kinodynamic path searcher to search for a safe tracking trajectory in the dense environments based on the target motion prediction.
	The trajectory is finally optimized to be safe and dynamically feasible. The proposed methods are integrated into an onboard quadrotor system with many engineering considerations.
	Abundant experiments and benchmark comparisons confirm the efficiency, effectiveness, and robustness of the proposed system.      
	\par
	In the future, we will focus on enabling dynamic obstacles avoidance during the tracking mission. 
	Moreover, we will improve the object detection method to make the proposed system applicable to more generalized scenarios.

	\newlength{\bibitemsep}\setlength{\bibitemsep}{0.0\baselineskip}
	\newlength{\bibparskip}\setlength{\bibparskip}{0pt}
	\let\oldthebibliography\thebibliography
	\renewcommand\thebibliography[1]{%
		\oldthebibliography{#1}%
		\setlength{\parskip}{\bibitemsep}%
		\setlength{\itemsep}{\bibparskip}%
	}
	\bibliography{icra2021_hzp}
\end{document}